\PassOptionsToPackage{table, x11names}{xcolor}

\documentclass[10pt,twocolumn,letterpaper]{article}

\usepackage[pagenumbers]{cvpr} 

\usepackage{amsmath,amsfonts,bm}

\def\rvf{{\mathbf{f}}}

\def\rmI{{\mathbf{I}}}

\def\vc{{\bm{c}}}

\DeclareMathAlphabet{\mathsfit}{\encodingdefault}{\sfdefault}{m}{sl}
\SetMathAlphabet{\mathsfit}{bold}{\encodingdefault}{\sfdefault}{bx}{n}

\DeclareMathOperator*{\argmin}{arg\,min}

\newcommand{\cb}{{\boldsymbol c}}

\newcommand{\xb}{{\boldsymbol x}}

\newcommand{\zb}{{\boldsymbol z}}

\newcommand{\x}{{\boldsymbol x}}

\newcommand{\z}{{\boldsymbol z}}

\newcommand{\epsilonb}{{\boldsymbol \epsilon}}

\newcommand{\Rd}{{\mathbb R}}

\newcommand{\Dc}{{\mathcal D}}
\newcommand{\Ec}{{\mathcal E}}

\newcommand{\Nc}{{\mathcal N}}
\newcommand{\Mc}{{\mathcal M}}

\newcommand{\Tc}{{\mathcal T}}
\newcommand{\Lc}{{\mathcal L}}

\usepackage{amsmath}

\usepackage{algorithm}
\usepackage{algpseudocode}

\usepackage[utf8]{inputenc} 
\usepackage[T1]{fontenc}    
\usepackage{url}            
\usepackage{booktabs}       
\usepackage{amsfonts}       
\usepackage{nicefrac}       
\usepackage{microtype}      
\usepackage{xcolor}       
\usepackage{xkcdcolors}
\usepackage{times}
\usepackage{epsfig}
\usepackage{graphicx}
\usepackage{placeins}
\usepackage{multirow}
\usepackage{caption}
\usepackage[skip=5pt]{caption}
\usepackage{rotating}
\usepackage{cancel}
\usepackage{setspace}

\usepackage{bm}
\usepackage{bibunits} 

\usepackage{amssymb,amsthm}
\usepackage{amsmath}
\usepackage{graphicx}
\usepackage{wrapfig,lipsum,booktabs}

\usepackage{epsfig}
\usepackage{tikz}
\usetikzlibrary{spy}
\usepackage{algpseudocode}
\usepackage{algorithm}
\usepackage{mathrsfs}

\usepackage{nicefrac}       
\usepackage{booktabs}       

\usepackage{thmtools,thm-restate}
\usepackage{listings}
\usepackage{lstautogobble}  %
\usepackage{color}          %
\usepackage{zi4}

\usepackage{indentfirst}

\usepackage{makecell}
\usepackage{tabularx}
\usepackage{animate}
\usepackage{capt-of}

\usepackage{colortbl} 
\usepackage[table]{xcolor}

\graphicspath{{small_figs/}}
\definecolor{cvprblue}{rgb}{0.21,0.49,0.74}
\definecolor{mylightgray}{RGB}{236, 236, 236}
\usepackage[pagebackref,breaklinks,colorlinks,allcolors=cvprblue]{hyperref}

\renewcommand{\sectionautorefname}{Sec.}

\def\paperID{7165} 
\def\confName{CVPR}
\def\confYear{2025}

\newcommand{\add}[1] {\textcolor{black}{#1}} 

\title{Optical-Flow Guided Prompt Optimization for Coherent Video Generation}

\author{Hyelin Nam$^{*, 1, 2}$, Jaemin Kim$^{*, 1}$, Dohun Lee$^{1}$, Jong Chul Ye$^{1}$ \\
Kim Jaechul Graduate School of AI, KAIST$^1$,  EverEx$^2$\\
{\footnotesize $^*$: Equal Contribution}\\
\tt\small{hyelin.nam.028@gmail.com,}
\tt\small{\{kjm981995, leedh7, jong.ye\}@kaist.ac.kr}
}

\begin{document}

\twocolumn[{%
\renewcommand\twocolumn[1][]{#1}%
\maketitle
\vspace{-0.7cm}
\begin{center}
    \newcommand{\numColumns}{4}
    \newcommand{\columnSpacing}{0.1cm}
    \begin{tabularx}{\textwidth}{XXXX}
        \centering \small \textbf{Base} & 
        \centering \small \textbf{Ours} & 
        \centering \small \textbf{Base} &
        \centering \small \textbf{Ours}
    \end{tabularx}
    \begin{tabular}{
        @{}
        p{\dimexpr(\textwidth-\columnSpacing*(\numColumns-1))/\numColumns} @{\hspace{\columnSpacing}}
        p{\dimexpr(\textwidth-\columnSpacing*(\numColumns-1))/\numColumns} @{\hspace{\columnSpacing}}
        p{\dimexpr(\textwidth-\columnSpacing*(\numColumns-1))/\numColumns} @{\hspace{\columnSpacing}}
        p{\dimexpr(\textwidth-\columnSpacing*(\numColumns-1))/\numColumns} @{}
    }
        \animategraphics[loop, width=\linewidth]{8}{small_videos/1_1/}{00}{15} &
        \animategraphics[loop, width=\linewidth]{8}{small_videos/1_2/}{00}{15} &
        \animategraphics[loop, width=\linewidth]{8}{small_videos/1_3/}{00}{15} &
        \animategraphics[loop, width=\linewidth]{8}{small_videos/1_4/}{00}{15}
    \end{tabular}
    \begin{tabularx}{\textwidth}{XX}
        \centering \small {\fontfamily{phv}\selectfont "Close up video of japanese food."} & 
        \centering \small {\fontfamily{phv}\selectfont "Person dancing in a dark room."}
    \end{tabularx}

    \vspace{0.5em}
    \begin{tabularx}{\textwidth}{XXXX}
        \centering \small \textbf{Base} & 
        \centering \small \textbf{Ours} & 
        \centering \small \textbf{Base} &
        \centering \small \textbf{Ours}
    \end{tabularx}
    \renewcommand{\numColumns}{4}
    \renewcommand{\columnSpacing}{0.25em}
    \begin{tabular}{
        @{}
        p{\dimexpr(\textwidth-\columnSpacing*(\numColumns-1))/\numColumns} @{\hspace{\columnSpacing}}
        p{\dimexpr(\textwidth-\columnSpacing*(\numColumns-1))/\numColumns} @{\hspace{\columnSpacing}}
        p{\dimexpr(\textwidth-\columnSpacing*(\numColumns-1))/\numColumns} @{\hspace{\columnSpacing}}
        p{\dimexpr(\textwidth-\columnSpacing*(\numColumns-1))/\numColumns} @{}
    }
        \animategraphics[loop, width=\linewidth]{8}{small_videos/2_1/}{00}{15} &
        \animategraphics[loop, width=\linewidth]{8}{small_videos/2_2/}{00}{15} &
        \animategraphics[loop, width=\linewidth]{8}{small_videos/2_3/}{00}{15} &
        \animategraphics[loop, width=\linewidth]{8}{small_videos/2_4/}{00}{15}
    \end{tabular}
    \begin{tabularx}{\textwidth}{XX}
        \centering \small {\fontfamily{phv}\selectfont "Time lapse video of a farm during sunset."} & 
        \centering \small {\fontfamily{phv}\selectfont "Video of a train tracks."}
    \end{tabularx}
    
    \vspace*{0.2cm}
    
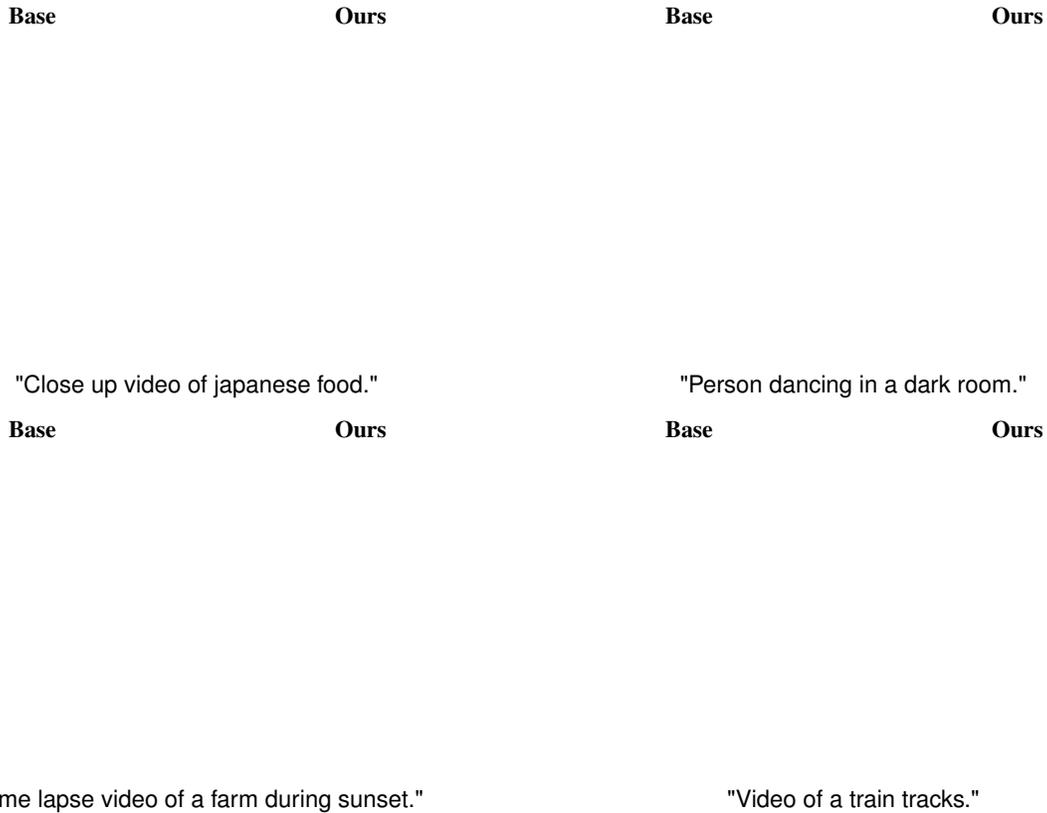
\captionof{figure}{
    \textbf{MotionPrompt} enhances temporal consistency and motion smoothness in text-to-video diffusion models by combining optical flow guidance with prompt optimization. It can be combined with a range of text-to-video diffusion models to produce visually coherent video sequences that closely align with intended motion while preserving content fidelity. \textit{Best viewed with Acrobat Reader. Click each image to play the video clip.}}
    \label{fig:main}
\end{center}
}]

\begin{abstract} 
While text-to-video diffusion models have made significant strides, many still face challenges in generating videos with temporal consistency. Within diffusion frameworks, guidance techniques have proven effective in enhancing output quality during inference; however, applying these methods to video diffusion models introduces additional complexity of handling computations across entire sequences.To address this, we propose a novel framework called \textbf{MotionPrompt} that guides the video generation process via optical flow. Specifically, we train a discriminator to distinguish optical flow between random pairs of frames from real videos and generated ones. Given that prompts can influence the entire video, we optimize learnable token embeddings during reverse sampling steps by using gradients from a trained discriminator applied to random frame pairs. This approach allows our method to generate visually coherent video sequences that closely reflect natural motion dynamics, without compromising the fidelity of the generated content.
We demonstrate the effectiveness of our approach across various models. Project Page: \url{https://motionprompt.github.io/}
\end{abstract}

\section{Introduction}
\label{sec:intro}
Recently, diffusion models have become the de-facto standard for image generation. In particular, text-to-image (T2I) diffusion models~\cite{dhariwal2021diffusion, rombach2022high} have gained significant attention due to their ability to enable users control over the generation process via text prompts. 
Building on these advancements, research has now progressed toward text-to-video (T2V) diffusion models~\cite{chen2024videocrafter2, guo2023animatediff, wang2023lavie} that generate videos which are both visually engaging and contextually coherent. Despite these advances, achieving temporally consistent videos remains a challenge in many T2V models. \add{Although recent work has attempted to address this, many proposed methods rely on fine-tuning~\cite{ge2023preserve} or introduce unnecessary interventions that can potentially adversely impact~\cite{chen2024unictrl, lee2024videoguide}}.

In diffusion frameworks, guidance is a commonly used technique to obtain samples aligned with specific objectives during inference. For example, in diffusion inverse solvers (DIS)~\cite{chung2023diffusion,song2023pseudoinverseguided,kim2024dreamsampler}, the guidance is usually given in the form of the likelihood function from the data consistency terms. However, a key technical issue in diffusion model guidance is that intermediate samples are corrupted by the additive Gaussian noise from the forward diffusion processes, making the computation of the likelihood term computationally prohibitive~\cite{chung2023diffusion}. To address this issue, diffusion posterior sampling (DPS) \cite{chung2023diffusion} approximates the likelihood function around the posterior mean, computed using Tweedie's formula~\cite{efron2011tweedie}. Another way to applying guidance is fine-tuning of the underlying diffusion models. For example, in DiffusionCLIP~\cite{kim2022diffusionclip}, the diffusion model for reverse sampling is fine-tuned using a directional CLIP guidance to generate images that align with a target text prompt.

However, applying these well-established techniques to video diffusion models (VDM) poses significant technical challenges. Unlike image generation, VDMs require the modeling of dependencies of across frames. For instance, if DPS is applied to VDMs, backpropagation must occur across all frames, making the process not only computationally expensive but also prone to instability. Applying fine-tuning of a VDM is even more challenging due to the model's large size. As a result, there remains a lack of a reliable, computationally efficient guidance mechanism that ensures temporal coherence across frames-a crucial factor in generating realistic videos.

Recently, semantic-preserving prompt optimization method has been proposed for generating minority images~\cite{um2024minorityprompt}. This approach integrates learnable tokens into a given prompt $\textit{P}$ and updates their embeddings on-the-fly during inference based on specific optimization criteria. 
While the original motivation of this work was different from reducing computational burden, we found that this on-the-fly prompt optimization concept aligns well with VDM guidance as a novel and computationally efficient guidance method, as the text prompt can influence all frames simultaneously.

Specifically, our new method, \textit{MotionPrompt}, is designed to enhance temporal consistency in video generation  while preserving the semantics through inference-time prompt optimization. By leveraging the global influence of the text prompt, MotionPrompt reduces the computational demands of entire sequence guidance, enabling indirect control over the latent video representation through gradients computed from only a subset of frames.  Specifically, we append a placeholder string $\textit{S}$ to the prompt $\textit{P}$ similar to \cite{um2024minorityprompt}, which serves as a marker for the learnable tokens. This approach preserves desired attributes across the video sequence and maintain the semantic meaning of the original prompt.

To further enhance temporal coherence in generated videos while preserving semantic integrity, our method incorporates a discriminator that uses optical flow to evaluate temporal consistency and guide prompt optimization. Specifically, we first train a discriminator to distinguish optical flow between random pairs of frames from real and generated videos. During sampling, a subset of generated frames is evaluated by the discriminator to assess the realism of their relative motion, as measured by optical flow. This process enables MotionPrompt to refine generated frames, achieving more natural and realistic motion patterns.
In summary, our contributions are as follows:
\begin{itemize}
    \item We propose MotionPrompt-a novel video guidance method that uses on-the-fly semantic prompt optimization to enhance temporal consistency and motion coherence in generated videos, without requiring diffusion models retraining or gradient calculations for every frame.
    \item By utilizing an optical flow-based discriminator to guide prompt optimization, we enforce temporal consistency in generated videos, enabling smoother, more realistic motion by aligning flows with real-world patterns \add{while minimizing impact on samples already close to real videos}.
\end{itemize}


\section{Related Works}
\label{sec:related_work}
\noindent\textbf{Latent Video Diffusion Models.} Latent video diffusion models~\cite{he2022latent, hong2023cogvideo, yang2024cogvideox, blattmann2023align} have gained attraction for their ability to efficiently generate videos by operating within a compressed latent space, thereby reducing the computational cost and memory demands associated with high-resolution video generation. Building on this approach, VideoCrafter~\cite{chen2023videocrafter1, chen2024videocrafter2}, AnimatedDiff~\cite{guo2023animatediff} and Lavie~\cite{wang2023lavie} extended latent diffusion to handle text-to-video generation to produce contextually relevant and controllable video outputs. While these advancements have significantly improved video generation, ensuring temporal consistency remains challenging. FreeInit~\cite{wu2023freeinit} alleviates this problem by refining the initial noise to ensure low-frequency information. While temporal consistency is increased, the process of iteratively obtaining clean videos is computationally expensive and can result in a loss of detail. \add{UniCtrl~\cite{chen2024unictrl} and VideoGuide~\cite{lee2024videoguide} address this issue by introducing attention injection and leveraging different pre-trained video diffusion models, respectively. However, these approaches can sometimes negatively impact performance: UniCtrl's injection mechanism struggles with handle color changes, while VideoGuide depends on the performance of the additional pre-trained model.}
Nevertheless, these methods are orthogonal to ours, as we optimize the prompt, and can be combined to further enhance video generation quality.
\begin{figure*}[t]
    \centering
    \includegraphics[width=1.0\linewidth]
    {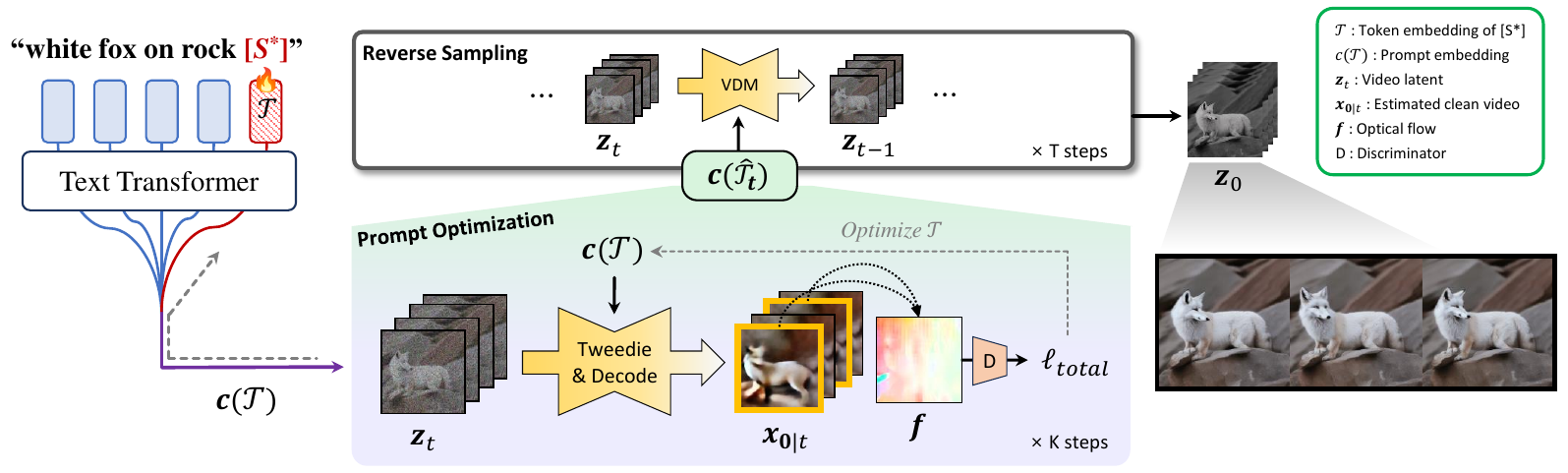}
    \caption{\textbf{Overall pipeline of MotionPrompt}. MotionPrompt enhances temporal consistency in text-to-video diffusion models by combining prompt optimization with an optical flow-based discriminator. Leveraging gradients from a subset of frames and aligning optical flow with real-world motion patterns, MotionPrompt efficiently generates videos with smooth, realistic motion and strong contextual coherence.}
    \vspace{-0.2cm}
    \label{fig:pipeline}
\end{figure*}

\noindent\textbf{Video Synthesis with Optical Flow.} Optical flow is a vector field that captures pixel-level motion patterns between two images. It has long been used to extract and control motion in visual data~\cite{siarohin2019first, siarohin2019animating, geng2024motion}. In the context of video synthesis, optical flow is particularly useful for guiding the generation of coherent and smooth motion. In image-to-video generation, several methods bypass the challenge of directly producing videos with natural motion by first generating optical flow and then using it as a basis for video synthesis~\cite{li2018flow, pan2019video, ni2023conditional}. Similarly in video inpainting, optical flow completion models are used prior to full video inpainting to produce natural and temporally consistent videos~\cite{gu2023flowguided, lee2024inpainter}. FlowVid~\cite{liang2024flowvid} trained a video diffusion models (VDM) that leverages optical flow as a temporal cue to achieve consistent video-to-video synthesis. Inspired by these, we hypothesize that improving the optical flow of the generated video would directly enhance its temporal consistency and motion smoothness in T2V generation with VDMs. Unlike previous approaches that use a separate optical flow generation module or reference optical flow, we employ a discriminator trained to assess the realism of the input optical flow and to guide sampling towards more realistic optical flow. 

\noindent\textbf{Prompt Optimization.} Prompting is an effective inference-time technique for enhancing pre-trained model performance on specific subtasks by guiding model responses with carefully crafted inputs. Widely used in modern language and vision-language models~\cite{wei2022chain, zhou2022coop, zhou2022cocoop}, this concept has also been extended to text-to-image diffusion models, proving effective across diverse tasks such as image editing~\cite{wen2023hard, park2024energy}, minority sampling~\cite{um2024minorityprompt}, and inverse problem solving~\cite{chung2024prompt}. However, these works mainly focus on image domain, and prompt optimization remains under-explored in the context of video generation. To the best of our knowledge, this is the first work to apply on-the-fly prompt tuning to VDMs. \\

\section{Main Contribution: MotionPrompt}

\subsection{Conditional Video Diffusion Model}
\label{sec:Prelim}
Latent video diffusion models encode $N$-frame clean video, $\{\x^{(i)}\}_{i=1}^N \in \Rd^{N \times C \times H \times W}$, to $\{\z_0^{(i)}\}_{i=1}^N$ using an encoder $\mathcal{E}$, where $C,H$ and $W$ represent channel, height and width of the video, respectively. Unless otherwise noted, we simply denote $\z_0$ as $\{\z_0^{(i)}\}_{i=1}^N$ for convenience
and $\z_0\sim p_0(\z)$.

The diffusion model aims to estimate the noise in the noised latent $\z_t$ from the forward diffusion process~\cite{dhariwal2021diffusion}
\begin{equation}
\label{eq:forward}
    q(\z_t | \z_0) = \mathcal{N}(\z_t; \sqrt{1 - \Bar{\alpha}_t} \, \z_{0}, \Bar{\alpha}_t \mathbf{I}),
\end{equation}
where $\Bar{\alpha}_t$ is the noise scheduling coefficient at timestep $t$. In the text-to-video diffusion model, the text condition is provided as an additional input. Given a prompt $\textit{P}$, the text embedding $\mathbf{c}$ is obtained through the text encoder $\mathcal{E}_{\text{text}}$, i.e., $\mathbf{c} = \mathcal{E}_{\text{text}}(\textit{P})$. 
Then, the training objective is to minimize
\begin{equation}
\label{eq:denosing_objective}
    \mathbb{E}_{\mathbf{z}_0, \bm{\epsilon}, t, \mathbf{c}} \left[ \|\bm{\epsilon} - \bm{\epsilon}_\theta(\mathbf{z}_t, t, \mathbf{c})\|^2 \right],
\end{equation}
where $\bm{\epsilon} \sim \mathcal{N}(0, \mathbf{I})$, and
$\bm{\epsilon}_\theta(\mathbf{z}_t, t, \mathbf{c})$ denotes the diffusion model parameterized by
$\theta$ with the text condition $\mathbf{c}$ and the noisy latent $\mathbf{z}_t$ at $t$.
Once the diffusion model is trained,  reverse diffusion sampling is performed.
For example, in DDIM~\cite{song2020denoising}, the reverse diffusion follows:
\begin{align}
\label{eq:reverse_sampling}
    \z_{t-1} &= \sqrt{\Bar{\alpha}_{t-1}} \hat{\z}_{t}  + \sqrt{1 - \Bar{\alpha}_{t-1}} \bm{\epsilon}_\theta(\z_t, t, \vc) \\
    \label{eq:tweedie}
    \hat{\z}_{t} &= \frac{1}{\sqrt{\Bar{\alpha}_t}} \left( \z_t - \sqrt{1 - \Bar{\alpha}_t} \bm{\epsilon}_\theta(\z_t, t, \vc) \right),
\end{align}
where  $\hat{\z}_{t}$ denotes the denoised sample at $t$, is obtained from Tweedie's formula~\cite{efron2011tweedie}. To enhance the impact of the text condition, we applied classifier-free guidance (CFG)~\cite{ho2021classifierfree} to all $\bm{\epsilon}$ predictions. The modified prediction is given by:
\begin{equation}
    \bm{\epsilon}^{w}_\theta(\mathbf{z}_t, t, \mathbf{c}) = \bm{\epsilon}_\theta(\mathbf{z}_t, t, \emptyset) + w \left[\bm{\epsilon}_\theta(\mathbf{z}_t, t, \mathbf{c}) - \bm{\epsilon}_\theta(\mathbf{z}_t, t, \emptyset)\right],
\label{eq:cfg}
\end{equation}
where $w$ denotes the guidance scale and $\emptyset$ represents the null text prompt. Thus, unless specified otherwise, $\bm{\epsilon}_\theta(\cdot)$ denotes terms with CFG applied.


So far, we have discussed text conditioned diffusion sampling. Moving beyond this, to navigate the sampling process in a way that minimizes a general  loss function $\ell(\z)$, it is essential to find the solution on the correct clean manifold: 
 \begin{align}
 \min_{\zb\in \Mc} \ell(\zb)
 \label{eq:opt}
 \end{align}
 where $\Mc$ represents the clean data manifold sampled from the unconditional distribution $p_0(\z)$.
In DPS \cite{chung2023diffusion}, this is achieved by 
enforcing the updated estimate from the noisy sample $\x_t \in \Mc_t$  to be constrained to stay on the same noisy manifold $\Mc_t$. 
\begin{equation}
\begin{aligned}
\label{eq:guidance}
    \z_{t-1} &= \sqrt{\bar\alpha_{t-1}}\left(\z_t -  \gamma_t \nabla_{\z_t}\ell (\hat\z_t)\right)+ \sqrt{1-\bar\alpha_{t-1}} \bm{\epsilon}_\theta(\z_t, t, \vc),
\end{aligned}
\end{equation}
where $\gamma_t>0$ denotes the step size. 

\subsection{Prompt Optimization for Video Guidance}
\label{subsec:po}
 
While direct guidance of the latent representation using \eqref{eq:guidance} has proven effective in image generation, calculating $\nabla_{\z_t}\ell (\hat\z_t)$ in the video domain poses significant challenges. 
Specifically, calculating the gradient of all frames is computationally expensive. Providing guidance for only  selected frames may reduce memory usage, but this can disrupt frame-to-frame consistency, resulting in inconsistencies in appearance, motion, and coherence throughout the video. 



To address this, we employ the prompt optimization method
and extend it to capitalize the text prompt's influence across the entire video. This approach enables indirect control of the latent video representation by using gradients derived from only a subset of frames, rather than necessitating gradients for every frame.
Specifically, instead of using \eqref{eq:opt} for the latent, we introduce an inference-time optimization problem with respect to the text embedding $\cb$:
 \begin{align}
 \hat\cb_t = \arg\min_{\cb} \ell(\zb_t,\cb)
 \label{eq:opt_prompt}
 \end{align}
One of the most important advantages of this approach is that it enables the use of a simple reverse diffusion process:
\begin{align}
\label{eq:guidance1}
    \z_{t-1} &= \sqrt{\bar\alpha_{t-1}}\hat{\z}_t(\hat\cb_t) + \sqrt{1-\bar\alpha_{t-1}} \bm{\epsilon}_\theta(\z_t, t, \hat \cb_t) \\
     \hat{\z}_{t}(\hat\cb_t) & =  \left( \z_t - \sqrt{1 - \alpha_t} \bm{\epsilon}_\theta(\z_t, t, \hat\vc_t) \right)/{\sqrt{\Bar{\alpha}    _t}} \label{eq:mytweedie}
\end{align}

Furthermore, to preserve the semantic meaning of the original prompt, rather than optimizing the entire text embedding $\vc$, we follow the approach introduced in ~\citet{um2024minorityprompt}, attaching learnable token embeddings to the end of the prompt and optimizing only these embeddings. Specifically, we first add new text tokens $\textit{S} = \{S_i\}_{i=1}^{n}$ to the tokenizer vocabulary and initialize their embeddings with words that can help improve video quality. In this work, we use the word ‘authentic’ and similar terms.  We then append these learnable tokens to the end of the given text prompt (e.g., "White fox on the rock." $\rightarrow$ "White fox on the rock $S_1$ ... $S_n$.").
\add{We denote this modified prompt as $\textit{P} + \textit{S}$. This leads to the following modified optimization problem:
\begin{align}
    \hat\cb_{t}:= \cb(\hat\Tc_t),&\quad \hat\Tc_t = \argmin_{\Tc} \ell(\zb_t, \vc(\Tc)),
    \label{eq:po}
\end{align}
where $\Tc$ denotes the embeddings of tokens $\textit{S}$, and $\vc(\Tc) := \mathcal{E}_{\text{text}}(\textit{P} + \textit{S})$.} This optimization occurs at each timestep $t$ within the defined range, causing $\Tc$ to evolve over time and, consequently, making $\vc(\Tc)$ vary with each timestep. By preserving the other token embeddings in the original text prompt, we ensure that essential text information is retained without loss and ensure the diffusion sampling trajectory on the correct manifold.  After the specified range, we revert to the original prompt $\textit{P}$ to maintain the overall appearance and structure of the initial video.

\subsection{Defining  Loss Function from Optical Flow}
\label{sec:loss}
In this section, we introduce the objective function $\ell(\z)$ to ensure the temporal coherence in the generated video. The total objective function is formulated as follows:
\begin{align*}
    \ell_{\text{total}}(\z_t,  \Tc) := & \, \lambda_{1} \ell_{\text{disc}}(\z_t, \cb(\Tc)) + \lambda_{2} \ell_{\text{TV}}(\z_t, \cb(\Tc)) \\
    & + \lambda_{3} ||\Tc - \Tc_0||_2^2,
\end{align*}
where  $\lambda_{1}$, $\lambda_{2}$ and $\lambda_{3}$ are regularization parameters. 
The $l_2$ loss term in $\ell_{\text{total}}$ is to ensure that the optimized token embedding is not far from the original token embedding space.
The other two loss terms $\ell_{\text{disc}}$ and $\ell_{\text{TV}}$ will be explained in detail soon. See Algorithms for the pseudo-code of the generation processes with our prompt optimization. 
%


\begin{figure*}[t]
\centering
\begin{minipage}{0.48\textwidth}
    \begin{algorithm}[H]
    \small
    \setstretch{1.28}
    \caption{MotionPrompt}
    \begin{algorithmic}[1]
        \Require $\epsilonb_\theta, \textit{P}, \textit{T}, \Ec_{text}, \Dc, \text{TimeCond}(t) T$
        \State $\z_T \sim \Nc(0, \rmI)$
        \For{$t=T$ {\bfseries to} $1$}
            \If{$\text{TimeCond}(t)$}
                \State $\vc \gets \textcolor{Plum}{OptEmb} (\z_t, \epsilonb_\theta, \textit{P}+\textit{S}, \Ec_{text})$
            \Else{}
                \State $\vc \gets \Ec_{text}(\textit{P})$
            \EndIf
            \State $\epsilonb_t \gets$ CFG from \eqref{eq:cfg} 
            \State $\z_t \gets \text{Reverse sampling from \eqref{eq:reverse_sampling}}$
        \EndFor
        \State {\bfseries return} $\Dc(\z_0)$
    \label{alg:motionprompt}
    \end{algorithmic}
    \end{algorithm}
\end{minipage}
\vspace{-0.2cm}
\hfill
\begin{minipage}{0.50\textwidth}
    \setstretch{1.0}
    \small
    \begin{algorithm}[H]
    \caption{Prompt Optimization}
    \begin{algorithmic}[1]
        \Function{\textcolor{Plum}{OptEmb}}{$\z_t, \epsilonb_\theta, \textit{P}+\textit{S}, \Ec_{text}$}
            \For{$k = 1$ \textbf{to} $K$}
            \State $\vc(\Tc)  \gets \Ec_{text}(\textit{P}+\textit{S})$
            \State $\epsilonb_t(\vc(\Tc)) \gets$ CFG from \eqref{eq:cfg}
            \State $\hat{\zb}_t(\vc(\Tc)) \gets$ Tweedie's formula from \eqref{eq:tweedie}
            \State Select \& Decode frames: $\hat{\x}^1_t(\vc(\Tc))$, $\hat{\x}^2_t(\vc(\Tc))$
            \State $\mathbf{f} \gets OF(\hat{\x}^1_t(\vc(\Tc)),\hat{\x}^2_t(\vc(\Tc)))$
            \State $\ell_{total} \gets \lambda_{1}\ell_{disc}(\mathbf{f}) + \lambda_{2}\ell_{TV}(\mathbf{f})$
            \Statex \hspace{6em} $+ \lambda_{3}||\mathcal{T} - \mathcal{T}_0||_2^2$
            \State $\Tc \gets \Tc - \eta \nabla_{\Tc} \Lc_{total}$
            \EndFor
            \State $\hat{\vc} \gets \Ec_{text}(\textit{P}+\textit{S})$
            \State \textbf{return} $\hat\vc$
        \EndFunction
    \label{alg:optemb}
    \end{algorithmic}
    \end{algorithm}
\end{minipage}
\vspace{-0.2cm}
\end{figure*}

\noindent \textbf{Optical flow discriminator loss.}
The main idea of MotionPrompt is to utilize the realistic optical flow to guide the diffusion model to generate temporally
coherent video. 
Unfortunately, the primary challenge of utilizing optical flow in generation tasks lies in the inherent lack of paired supervised optical flow data. To address this, we aim to guide the sampling process toward aligning the optical flow of the generated video with that of real videos. 

Specifically, we employ a discriminator trained to distinguish between optical flow derived from generated videos and that from real ones. Note that by optimizing the prompt rather than the latent representation directly, we can design the optical flow discriminator to take a single flow as input, rather than requiring flow from entire video sequences. Specifically, given an optical flow model $OF(\x)$ that estimates the optical flow between two frames, the discriminator $\phi_d(\cdot): \Rd^{2\times H \times W} \rightarrow [0,1]$ is trained to minimize 
\begin{equation}
\label{disc_train_objective}
    \min_\theta-\mathbb{E}_{\mathbf{f}_r, \mathbf{f}_f} \left[ \log \phi_\theta(\mathbf{f}_r) + \log (1 - \phi_\theta(\mathbf{f}_f)) \right],
\end{equation}
where $\rvf_r:=OF(\x_r), \rvf_f:=OF(\x_f)$ are optical flows calculated between two selected frames from real and generated (fake) videos, $\x_r$ and $\x_f$, respectively. 
For further training details, please refer to the implementation section (\autoref{subsec:impl_detail}) and supplementary material.

Once the discriminator is trained, we use it to guide the video samples toward more realistic and temporally consistent outputs. Since the discriminator operates in the clean video frames, we use the denoised video obtained via Tweedie's formula, \add{defined as $\hat\xb_t(\cb(\Tc)):=\Dc(\hat{\z}_{t}(\cb_t(\Tc)))$ using \eqref{eq:mytweedie}. Then, we select a subset of frames and decode it. Let's assume we choose two frames, denoted as $\hat{\x}^1_{t}(\cb(\Tc))$ and $\hat{\x}^2_{t}(\cb(\Tc))$. While we exemplified with two frames, it is also possible to calculate the optical flow across more frames and use the mean of their individual losses as the total loss. 
Subsequently, we compute the optical flow between two frames $\rvf(\hat{\x}_t(\cb(\Tc))) :=  OF(\hat{\x}^1_{t}(\cb(\Tc)), \hat{\x}^2_{t}(\cb(\Tc)))$, and feed it into the trained discriminator. The sampling process is then adjusted to steer the output toward samples that the discriminator classifies as real:
\begin{equation}
\label{eq:disc_guide_objective}
    \ell_{\text{disc}}(\z_t, \cb(\Tc)) := \log (1 - \phi_{\theta^*}(\rvf(\hat{\x}_t(\cb(\Tc)))).
\end{equation}
where $\theta^*$ denotes the optimized discriminator weight.}

\noindent \textbf{TV loss for optical flow.}
Additionally, to ensure that the optical flow adheres to the assumption of a smooth field, we incorporate total variation (TV) loss as an additional regularization term. For simplicity, in the following equation, we will denote $\rvf(\hat{\x}_t(\cb(\Tc)))$ as $\mathbf{f}$. This TV loss is then defined as 
\begin{equation}
\label{eq:tv_objective}
    \ell_{TV}(\z_t, \vc(\mathcal{T})) := \sum_{i=1}^{H} \sum_{j=1}^{W} \left( |\mathbf{f}_{i,j} - \mathbf{f}_{i+1,j}| + |\mathbf{f}_{i,j} - \mathbf{f}_{i,j+1}| \right),
\end{equation}
where $\mathbf{f}_{i,j}$ is the value at the position ($i, j$) of $\mathbf{f}$.  

\section{Experiments}
\label{sec:Experiments}
\subsection{Implementation Details} 
\label{subsec:impl_detail}
\paragraph{Baselines.} To evaluate our method across different frameworks, we test it with open-source text-to-video diffusion models, including Lavie~\cite{wang2023lavie}, AnimateDiff~\cite{guo2023animatediff}, and Videocrafter2~\cite{chen2024videocrafter2}. For AnimateDiff, we use the \texttt{RealisticVision} pre-trained model\footnote{\url{https://civitai.com/models/4201?modelVersionId=29460}}. 
We generate videos using a DDIM sampler with 50 steps and 800 prompts from VBench~\cite{huang2023vbench}, ensuring both the baselines and our method use the same seed.\\

\begin{figure*}[t]
    \centering
    \includegraphics[width=1.0\linewidth]
    {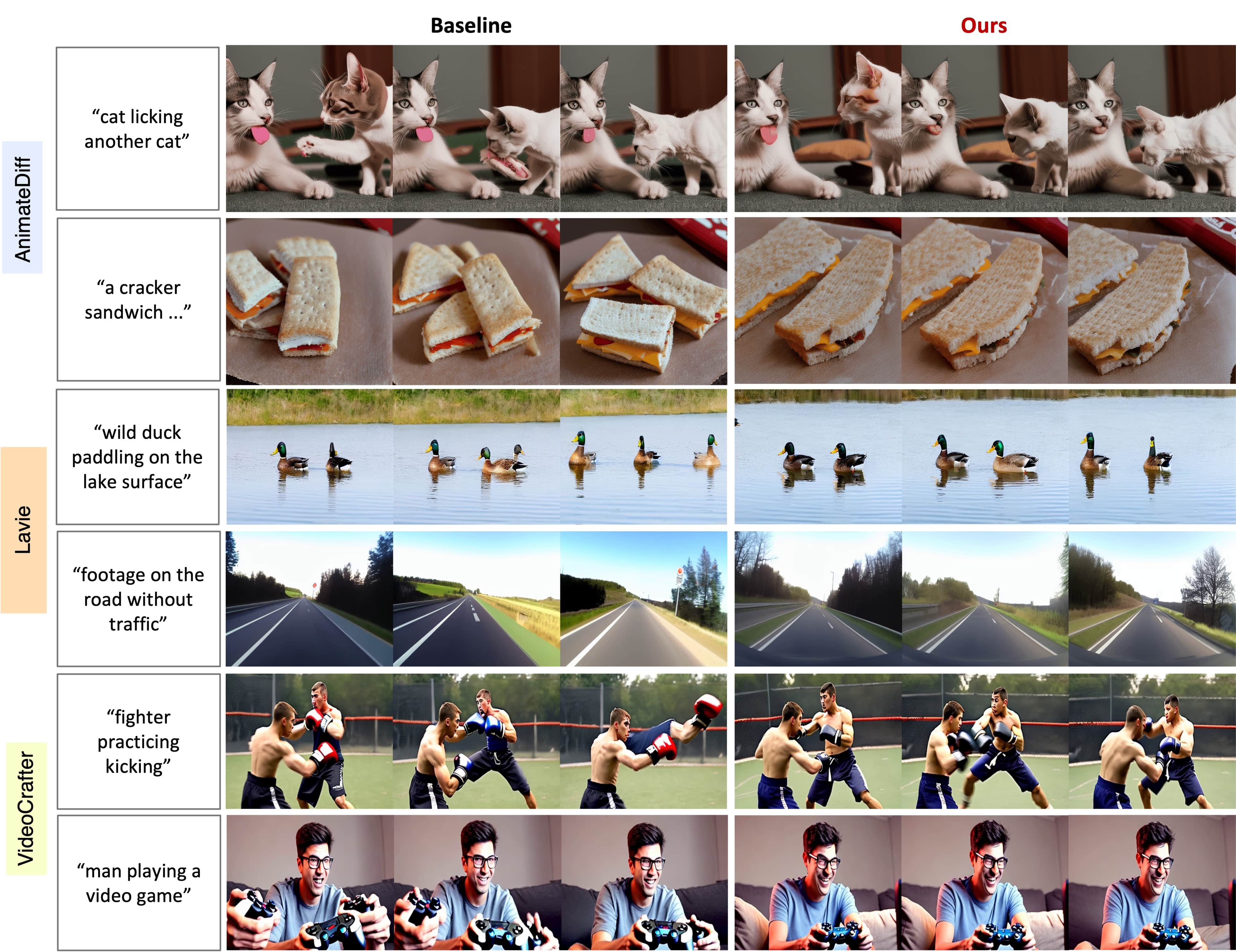}
    \caption{Qualitative comparison against three baselines. Additional results are provided in the supplementary material.}
    \vspace{-0.2cm}
    \label{fig:qualitative}
\end{figure*}

\noindent \textbf{Discriminator training.}
For discriminator training, we sample videos from each model using the same set of 800 prompts.
We highlight that the discriminator is trained on the optical flow of complete, clean videos, whereas during inference, it processes the optical flow of denoised estimates. This distinction ensures that, even with the same prompts, the discriminator encounters unseen data.
We fine-tune a pre-trained Vision Transformer (ViT)~\cite{dosovitskiy2021an} as an image encoder, adding a projection layer to adapt two-channel optical flow for ViT and using a 3-layer MLP as the classifier. This setup achieved rapid convergence in under 20 epochs. Optical flow was extracted with RAFT~\cite{teed2020raft} from real videos selected from the DAVIS~\cite{pont20172017} and WebVid~\cite{bain2021frozen}, as well as from generated videos. 
Finally, we train and deploy a separate discriminator for each video model. All training and experiments were conducted on a single 40GB NVIDIA A100 GPU.

\subsection{Results}
\begin{table*}
\centering
\small 
\resizebox{0.98\textwidth}{!}{%
\begin{tabularx}{\textwidth}{l>{\centering\arraybackslash}m{2cm}>{\centering\arraybackslash}m{2cm}>{\centering\arraybackslash}m{2cm}>{\centering\arraybackslash}m{2cm}>{\centering\arraybackslash}m{2cm}>{\centering\arraybackslash}m{2cm}}
\toprule
{} & \multicolumn{5}{c}{Temporal Quality} & \multicolumn{1}{c}{Text Alignment} \\
\cmidrule(lr){2-6} \cmidrule(lr){7-7}
Method & \makecell{Subject \\ Consistency ($\uparrow$)} & \makecell{Background \\ Consistency ($\uparrow$)} & \makecell{Temporal \\ Flickering ($\uparrow$)} & \makecell{Motion \\ Smoothness ($\uparrow$)} & 
\makecell{Dynamic \\ Degree ($\uparrow$)} &
\makecell{Overall \\ Consistency ($\uparrow$)} \\
\midrule
Lavie~\cite{wang2023lavie} & $0.9599$ & $0.9739$ & $0.9487$ & $0.9690$ & \textbf{0.5150} & $\textbf{0.2506}$ \\
Lavie + \textbf{Ours} & $\textbf{0.9646}$ & $\textbf{0.9781}$ & $\textbf{0.9625}$ & $\textbf{0.9765}$ & $0.3963$ & $0.2415$ \\
\midrule
AnimateDiff~\cite{guo2023animatediff} & $0.9488$ & $0.9755$ & $0.9228$ & $0.9578$ & $\textbf{0.4700}$ & $\textbf{0.2532}$ \\
AnimateDiff + \textbf{Ours} & $\textbf{0.9528}$ & $\textbf{0.9763}$ & $\textbf{0.9258}$ & $\textbf{0.9599}$ &  $0.4125$ & $0.2529$  \\
\midrule
VideoCrafter2~\cite{chen2024videocrafter2} & $0.9736$ & $0.9559$ & $0.9559$ & $\textbf{0.9750}$ & $\textbf{0.4088}$ & $\textbf{0.2498}$ \\
VideoCrafter2 + \textbf{Ours} & $\textbf{0.9745}$ & $\textbf{0.9774}$ & $\textbf{0.9588}$ & $0.9759$ & $0.3938$ & $0.2451$ \\
\bottomrule
\end{tabularx}
}
\caption{Quantitative evaluation of text-to-video generation. \textbf{Bold}: Best.}
\label{tab:results_t2v}
\vspace{-0.1cm}
\end{table*}

\paragraph{Qualitative comparisons.} We provide visual comparisons of our method against three baselines in \autoref{fig:qualitative}. The baselines struggle with maintaining temporal consistency, often failing to preserve the appearance or quantity of objects, and sometimes resulting in objects that suddenly appear or disappear. In contrast, the proposed framework effectively suppresses appearance changes and sudden shifts in video generation. Additionally, our method maintains a consistent color tone across all frames and accurately captures the scene attributes and details intended by the original prompts.

\paragraph{Quantitative comparisons.} For quantitative comparison, we evaluate five key metrics from VBench~\cite{huang2023vbench}—subject consistency, background consistency, temporal flickering, motion smoothness, and dynamic degree—to assess improvements in consistency and motion. We also measure overall consistency to confirm that prompt optimization maintains fidelity to the text prompt. \autoref{tab:results_t2v} shows that our method improves object consistency, reduces temporal flickering, and enhances motion smoothness with minimal impact on text alignment. \add{We are aware that there may exist a trade-off between consistency and motion dynamics in the proposed method. However,  visualization results  demonstrate that our method balances dynamics and coherence effectively.} See supplementary materials for more details.

\paragraph{User study.} 
Additionally, we conduct an A/B user study. We showed videos generated by the baseline and our method for the same prompts, asking participants to choose the one with better temporal quality.
\begin{wraptable}[4]{r}{0.4\linewidth}
    \vspace{-0.1cm}
    \centering
    \resizebox{\linewidth}{!}{
        \begin{tabular}{lccc} 
        \toprule
        Baseline & Win & Tie & Lose \\
        \midrule
        AnimateDiff & 66.5 & 17.8 & 15.7 \\
        Lavie & 55.1 & 21.1 & 23.8 \\
        VideoCrafter2 & 53.0 & 17.7 & 29.3 \\
        \bottomrule
        \end{tabular}
    }
    \caption{User study results.}
    \label{tab:userstudy}
\end{wraptable}
Among 100 participants evaluating a total of 90 videos (30 videos per model), our method was preferred overall (\autoref{tab:userstudy}).

\section{Additional Experiments}
\label{sec:Analysis}
We performed additional analysis to analyze the design components of the proposed approach and to support the effectiveness of our method, focusing on the AnimateDiff~\cite{wang2023lavie}.

\vspace{-0.1cm}

\subsection{Ablation Study}
\paragraph{Analysis of each loss component.} In ~\autoref{tab:hyperparameter}, we examine how key hyperparameters affect our framework’s performance. 
First, we validate the effectiveness and robustness of our loss components.
\textbf{(\boldmath{$\lambda_1$})} A comparison between rows (a) and (c) reveals that incorporating $\ell_{disc}$ further improves the results.  
(\boldmath{$\lambda_2$}) To investigate the role of $\lambda_2$, we gradually increased its value. As shown in rows (b)–(d), a higher weight results in smoother motion but comes at the cost of performance in other metrics.
Based on this observation, we selected a moderate weight to balance the trade-off.
(\boldmath{$\lambda_3$}) This loss was introduced after identifying a degradation in video quality, which we hypothesized was due to the optimized embedding deviating from the text space the diffusion model was trained on.
 Comparing rows (c) and (e) supports this hypothesis, as certain metrics show a decline. While performance varies with different hyperparameter choices, all configurations consistently outperform the baseline, demonstrating that our method remains robust and is not overly sensitive to hyperparameter selection.

\paragraph{Analysis of hyperparameters.} We also explored the optimal number of iterations per optimization step and the most effective stage within the 50-step sampling process.
For iteration count, a single iteration was insufficient, while 7 iterations smoothed motion but reduced dynamic degree and overall consistency, as well as increasing computation time.
An effective balance was achieved with 3 iterations. Similarly, starting optimization early improved motion smoothness but reduced consistency, while delaying it too late weakened its impact.
Based on these findings, we established an optimal range for applying optimization. Except for the hyperparameters being compared, all other settings use the optimal hyperparameters listed in the supplementary materials.

\begin{table}[t]
\centering
\resizebox{0.98\columnwidth}{!}{%
\begin{tabular}{ccccc}
\multicolumn{5}{c}{\textbf{\Large Loss Weight}} \\
\hline
 $\lambda_1$ / $\lambda_2$ / $\lambda_3$ & \makecell{Subject \\ Consistency ($\uparrow$)} & \makecell{Motion \\ Smoothness ($\uparrow$)}  & \makecell{Dynamic \\ Degree ($\uparrow$)} & \makecell{Overall \\ Consistency ($\uparrow$)}\\
\hline
(a) 0.0 / 5.0 / 10.0  & 0.9504 & 0.9643 & \textbf{0.4365} & \underline{0.2528} \\
(b) 1.0 / 0.0 / 10.0 & \underline{0.9522} & 0.9648 & \underline{0.4232} & 0.2522 \\
\rowcolor{mylightgray} 
(c) 1.0 / 5.0 / 10.0 & \textbf{0.9528} & \underline{0.9599} & 0.4125 & \textbf{0.2529} \\
(d) 1.0 / 10.0 / 10.0 & 0.9514 & \textbf{0.9658} & 0.4072 & 0.2525 \\
(e) 1.0 / 5.0 / 0.0 & 0.9505 & 0.9650 & \underline{0.4232} & 0.2517 \\
\end{tabular}
}
\vspace{0.3cm} 

\resizebox{0.85\columnwidth}{!}{%
\begin{tabular}{cccc}
\multicolumn{4}{c}{\textbf{Optimization Iterations}} \\ 
\hline
 iter & \makecell{Motion \\ Smoothness ($\uparrow$)} & \makecell{Dynamic \\ Degree ($\uparrow$)}  & \makecell{Overall \\ Consistency ($\uparrow$)} \\
\hline
1 & 0.9593 & \textbf{0.4550} & \underline{0.2527}  \\
\rowcolor{mylightgray} 
3 & \underline{0.9599} & \underline{0.4125} & \textbf{0.2529}  \\
7 & \textbf{0.9614} & 0.3950 & 0.2505 \\
\end{tabular}%
}
\vspace{0.5cm} 

\resizebox{0.95\columnwidth}{!}{%
\begin{tabular}{cccc}
\multicolumn{4}{c}{\textbf{Optimization Range}} \\ 
\hline
 timestep & \makecell{Motion \\ Smoothness ($\uparrow$)} & \makecell{Dynamic \\ Degree ($\uparrow$)}  & \makecell{Overall \\ Consistency ($\uparrow$)} \\
\hline
$t < 15$ & \textbf{0.9601} & 0.3600 & 0.2502 \\
\rowcolor{mylightgray} 
$3 < t < 15$ & \underline{0.9599} & \underline{0.4125} & \underline{0.2529} \\
$7 < t < 15$ & 0.9581 & \textbf{0.4173} & \textbf{0.2530} \\
\end{tabular}%
}

\caption{VBench metrics by hyperparameter. $t=0$ represents the inital noise. The highlighted row shows the final hyperparameter configuration, yielding well-balanced results. \textbf{Bold}: Best, \underline{Underline}: Second Best.}
\label{tab:hyperparameter}
\vspace{-0.3cm}
\end{table}

\begin{figure}[t]
    \centering
    \includegraphics[width=1.0\columnwidth]
    {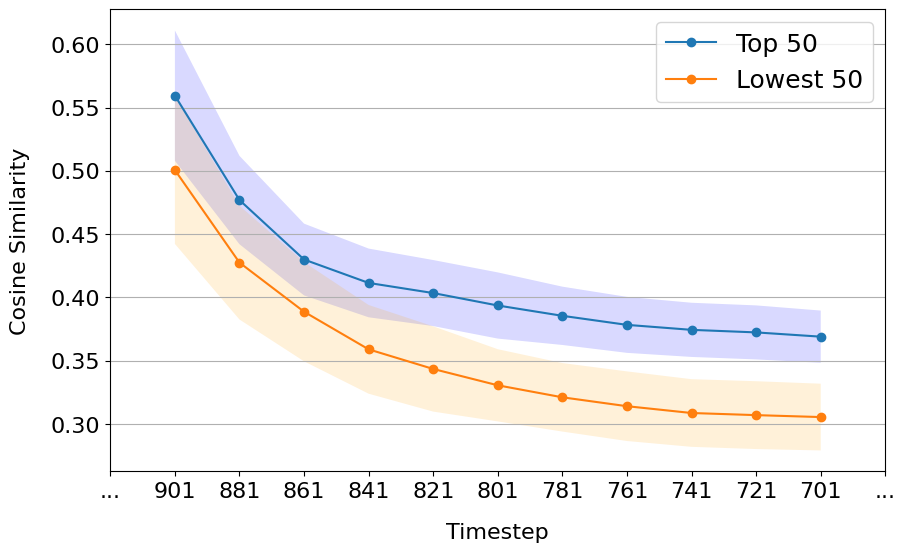}
    \caption{Cosine similarity between learnable and initial token embeddings. The cosine similarity decreases over time $t$, with more variation in embeddings observed for videos that initially exhibit lower subject consistency.}
    \vspace{-0.2cm}
    \label{fig:analysis_embeds}
\end{figure}


\subsection{Discriminator Generalization}
We primarily use a discriminator trained on paired data, where the model generating fake data matches the one used during inference.
To assess the discriminator's training impact and generalization, we also perform cross-dataset inference using a discriminator trained on a different dataset.
\autoref{tab:disc_gen} presents the results of this evaluation. 
Surprisingly, we observe a general improvement in performance when using a discriminator trained on different data (i.e., videos generated by Lavie and VideoCrafter2). 
We conjecture that this is because the baseline performance of Lavie and VideoCrafter2 is higher compared to AnimateDiff, making it more challenging for the discriminator to differentiate these videos from real ones.
This likely resulted in stricter training, which may have contributed to the improved performance observed during inference.
These findings suggest the potential for further performance enhancements through improved discriminator training strategies.

\begin{table}[t]
\centering
\resizebox{0.95\columnwidth}{!}{%
\begin{tabular}{c c c c}
\hline
 Source Model for Fake Data & \textbf{AD (Defualt)} & \textbf{Lavie} & \textbf{VC2} \\
\hline
Subject Consistency & 0.9528 & 0.9625 & 0.9535 \\
Background Consistency & 0.9763 & 0.9753 & 0.9764 \\
Temporal Flickering & 0.9258 & 0.9490 & 0.9283 \\
Motion Smoothness & 0.9599 & 0.9691 & 0.9617 \\
Dynamic Degree & 0.4125 & 0.4088 &  0.4100 \\
Overall Consistency & 0.2529 & 0.2473 & 0.2509 \\
\hline
\end{tabular}%
}
\caption{Quantitative results obtained using a discriminator trained on a different dataset. AD and VC2 denote AnimateDiff and VideoCrafter 2, respectively.}
\vspace{-0.4cm}
\label{tab:disc_gen}
\end{table}

\subsection{Token Variations}
To demonstrate that the improvement in temporal consistency is not simply due to the addition of prompt, we measure the cosine similarity between the initial token embedding $\mathcal{T}$ and the optimized embedding at each timestep $t$. As shown in \autoref{fig:analysis_embeds}, the cosine similarity decreases with sampling step and the rate of change gradually converging. Additionally, we compare the average cosine similarity using the top 50 videos and lowest 50 videos based on the subject consistency scores. We observe that videos with higher consistency exhibited less token variation, indicating that our method performs less optimization on videos which the discriminator judges to be closer to real video, thereby preserving their original characteristics.

\begin{figure}[t]
    \centering  \includegraphics[width=1.0\columnwidth]
    {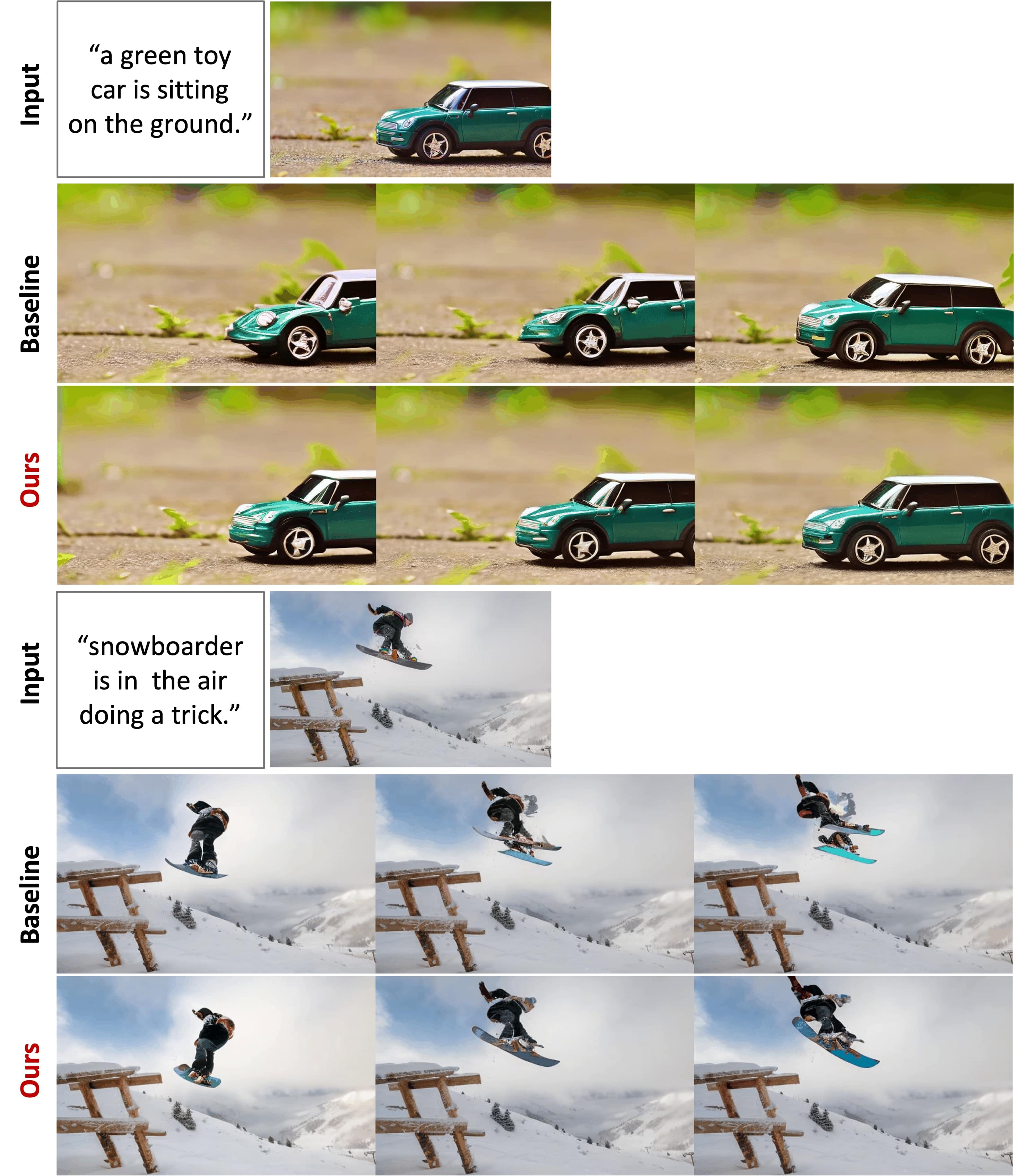}
    \caption{Comparison of video results generated by the vanilla DynamiCrafter model and our method.}
    \vspace{-0.5cm}
    \label{fig:analysis_dc}
\end{figure}

\subsection{Extensions: Image-to-Video Diffusion Model}  To further verify our framework's capabilities, we extended it to an image-to-video (I2V) diffusion model, DynamiCrafter~\cite{xing2025dynamicrafter}, which also uses text prompts as input. Although the vanilla model produced relatively consistent videos due to the reference image, issues arose with differences in appearance details and artifacts around objects. When combined with our method during sampling, these issues were significantly mitigated (\autoref{fig:analysis_dc}).


\section{Conclusion}
In this work, we introduced MotionPrompt, addressing the fundamental challenge in text-to-video models: generating temporally consistent and natural motion. Specifically, we leveraged optical flow, specifically using a discriminator trained to distinguish between optical flow from real videos and that from generated (i.e., fake) videos. By incorporating text optimization, our approach effectively addressed inefficiencies in guiding video models. Qualitative and quantitative experiments demonstrated the effectiveness of our proposed method.

\paragraph{Limitations.} While our approach improves baseline model results, it requires slightly more generation time. However, optimization is applied to only 10 to 15 steps, keeping costs low relative to performance gains. Additionally, since our objective function is not grounded in physics, improvements may not always produce physically plausible outcomes. We anticipate that as foundational models capable of assessing physical plausibility using representations extracted from video—such as optical flow or point correspondence—advance, this limitation can be mitigated.

\section*{Acknowledgement}
\noindent We appreciate the fruitful discussions with Soobin Um. This work was supported by the National Research Foundation of Korea (NRF) under Grants RS-2024-00336454 and RS-2023-00262527, and by the Institute for Information \& Communications Technology Planning \& Evaluation (IITP) grant funded by the Korea government (MSIT) (RS-2019-II190075, Artificial Intelligence Graduate School Program, KAIST).

{
    \small
    \bibliographystyle{ieeenat_fullname}
    \bibliography{main}
}

\clearpage
\setcounter{page}{1}
\maketitlesupplementary

\renewcommand{\thesection}{\Alph{section}}
\setcounter{section}{0} 
\renewcommand{\sectionautorefname}{Suppl Sec.}

The supplementary sections are organized as follows. \autoref{sup:detail} delves into the details on our training, inference configurations and evaluation setups. \autoref{sup:additional_analysis} presents additional experimental analyses. Finally, \autoref{sup:results} provides further results obtained from our framework.

\section{Implementation and Evaluation Details}
\label{sup:detail}

\subsection{Training Details of the Discriminator}
The discriminator was trained to distinguish between real and generated optical flow representations, aiming to enhance temporal consistency in video generation. For training, we use a batch size of 32 and the SGD optimizer with a learning rate of 0.0005 and a momentum of 0.9. The training process spans approximately 20 epochs, with the model parameters from the epoch achieving the best validation loss being selected.



\subsection{Hyperparameters for Evaluation}
\autoref{tab:hyperparams} lists the hyperparameters used for our quantitative evaluation. The `\# of frames' denotes the number of video frames decoded into pixels and used for optical flow calculation. Specifically, if the value is 6, this indicates that 3 sets of adjacent frames were sampled, resulting in 3 optical flows being used to compute the corresponding loss.

\begin{table}[h]
\centering
\resizebox{0.9\columnwidth}{!}{%
\begin{tabular}{c c c c}
\hline
 & \textbf{Lavie} & \textbf{AnimateDiff} & \textbf{VideoCrafter2} \\
\hline
\# of tokens & 1 & 1 & 1 \\
$K$ (\text{opt iter}) & 3 & 3 & 3 \\
\text{opt range} & 5<$t$<15 & 3<$t$<15 & 3<$t$<20 \\
$\text{\# of frames}$ & 6 & 2 & 6 \\
$\lambda_{1}$ & 1.0 & 1.0 & 1.0 \\
$\lambda_{2}$ & 1.0 & 5.0 & 5.0 \\
$\lambda_{3}$ & 10.0 & 10.0 & 3.0 \\
$\eta$ (lr) & 0.0005 & 0.005 & 0.001 \\
\hline
\end{tabular}%
}
\caption{Evaluation hyperparameters used for each model.}
\vspace{-0.2cm}
\label{tab:hyperparams}
\end{table}

\subsection{User study}
We conduct an A/B user study on Prolific, where 100 participants compared total 90 pairs of videos. Each participant was asked to choose the video they found more natural and smooth in motion by asking ``Choose the more natural and smoothly moving video.''. The answer options were: A, B, or Both.

\begin{figure}[t]
    \centering
    \includegraphics[width=1.0\columnwidth]
    {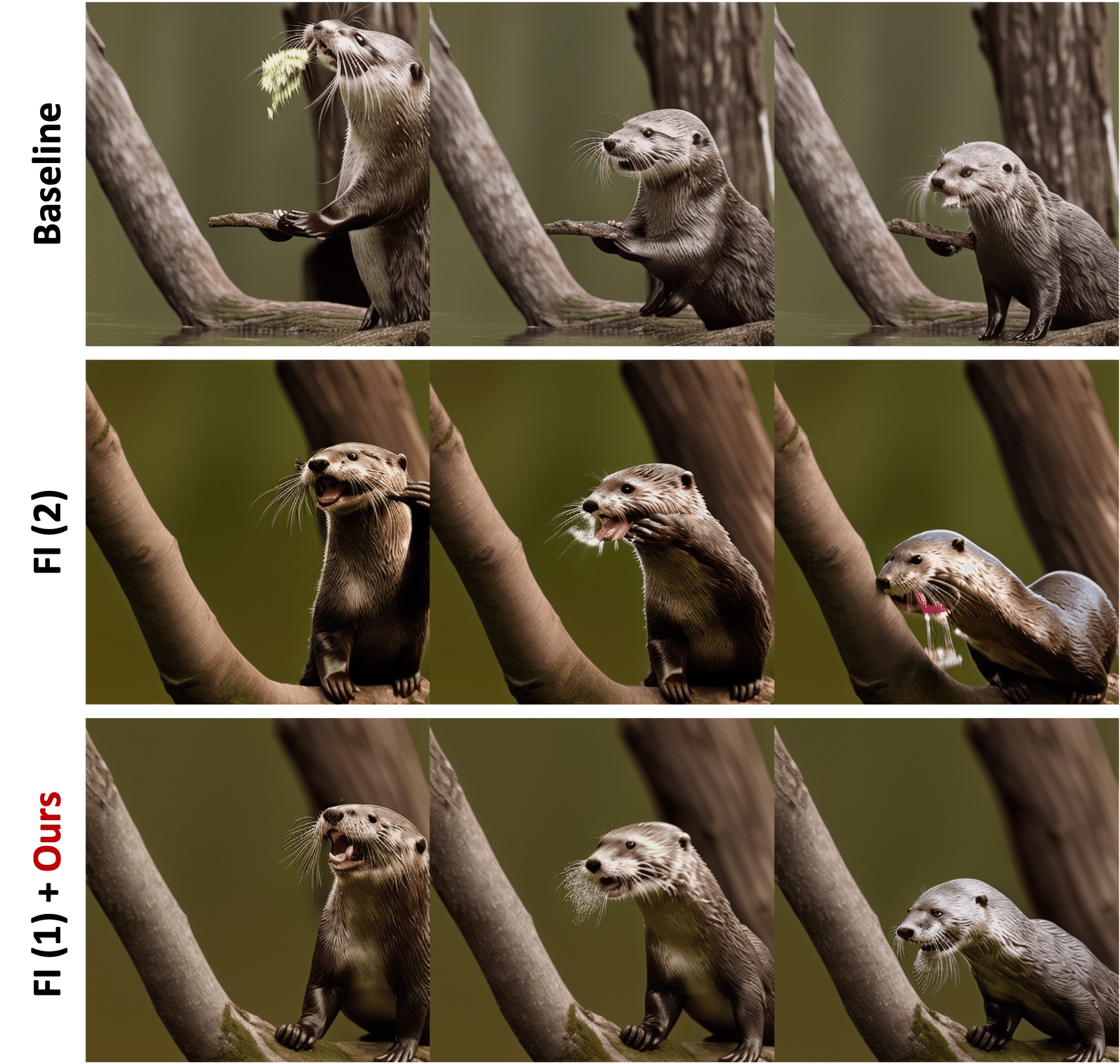}
    \caption{Qualitative comparison between the baseline, FreeInit, and FreeInit combined with our method. When FreeInit is used repeatedly, videos tend to lose detail and exhibit saturation issues. In contrast, combining our method with a single application of FreeInit mitigates these problems while improving temporal quality.}
    \vspace{-0.2cm}
    \label{fig:sp_freeinit}
\end{figure}

\section{Additional Analysis}
\label{sup:additional_analysis}

\begin{table*}[t]
\centering
\small 
\resizebox{0.95\textwidth}{!}{%
\begin{tabularx}{\textwidth}{l>{\centering\arraybackslash}m{2cm}>{\centering\arraybackslash}m{2cm}>{\centering\arraybackslash}m{2cm}>{\centering\arraybackslash}m{2cm}>{\centering\arraybackslash}m{2cm}>{\centering\arraybackslash}m{2cm}}
\toprule
{} & \multicolumn{5}{c}{Temporal Quality} & \multicolumn{1}{c}{Text Alignment} \\
\cmidrule(lr){2-6} \cmidrule(lr){7-7}
Method & \makecell{Subject \\ Consistency ($\uparrow$)} & \makecell{Background \\ Consistency ($\uparrow$)} & \makecell{Temporal \\ Flickering ($\uparrow$)} & \makecell{Motion \\ Smoothness ($\uparrow$)} & 
\makecell{Dynamic \\ Degree ($\uparrow$)} &
\makecell{Overall \\ Consistency ($\uparrow$)} \\
\midrule
Baseline & 0.9488 & 0.9755 & 0.9228 & 0.9578 & 0.4700 & 0.2532 \\
Baseline+ \textbf{Ours} & 0.9528 & 0.9763 & 0.9258 & 0.9599 & 0.4125 & 0.2529 \\
\midrule
Increased Tokens (3)& 0.9530 & 0.9760 & 0.9259 & 0.9605 & 0.3938 & $0.2498$ \\
Tokens Placed at Front & 0.9507 & 0.9730 & 0.9244 & 0.9615 & 0.3730 & 0.2423 \\
Init with `the' & 0.9509 & 0.9760 & 0.9263 & 0.9599 & 0.4438 & 0.2527 \\
\bottomrule
\end{tabularx}
}
\caption{Ablation results comparing the baseline, default setting, increased token count (3 tokens), tokens placed at the front, and tokens initialized with the word `the'. Evaluation metrics are reported for subject consistency, background consistency, temporal flickering, motion smoothness, dynamic degree, and overall consistency.}
\label{tab:sp_token_ablation}
\vspace{-0.1cm}
\end{table*}

In this section, we present additional analyses to evaluate the generalization capability and performance of our approach across various scenarios. The experiments primarily focus on AnimateDiff~\cite{guo2023animatediff}.


\subsection{Ablation on Token Optimization}
Here, we conduct an ablation study to investigate the impact of various configurations related to token optimization on the overall performance (\autoref{tab:sp_token_ablation}). Except for the factors being examined, all other settings were kept consistent with the values in \autoref{tab:hyperparams}.

\paragraph{The number of tokens}
First, we examine whether increasing the number of tokens enhances the optimization effect. Specifically, we added three additional tokens initialized as "authentic", "real" and "clear". Although this increased the factors that could potentially improve temporal quality, leading to an improvement in some temporal quality metrics, it also resulted in a decline in dynamic degree and overall consistency. Consequently, we determined that the benefits were not significant enough and chose to use a single token as the default setting.

\paragraph{The placement of tokens}
In addition, we investigate the effect of the learnable token's placement. Instead of appending the learnable token to the end of the given prompt, we placed it at the front of the prompt to evaluate its impact. Similar to the observations in ~\citet{um2024minorityprompt}, we also find that appending the token to the end of the given prompt is more effective. This aligns with the general practice of structuring sentences where content-related information is stated first, followed by descriptive elements like adjectives.

\paragraph{Robustness to Initialization Words}
In main paper, we demonstrate that the cosine similarity with the initialization token starts sufficiently low and gradually decreases over time, indicating that the improvement is not merely a result of adding tokens. To further support this, we initialize the token with a seemingly unrelated word, `the', and assess its impact on video generation quality. While the performance improvement was smaller compared to our final configuration, it still outperformed the baseline. This further reinforces the effectiveness and robustness of our method.

\begin{figure}[t]
    \centering
    \includegraphics[width=1.0\linewidth]
    {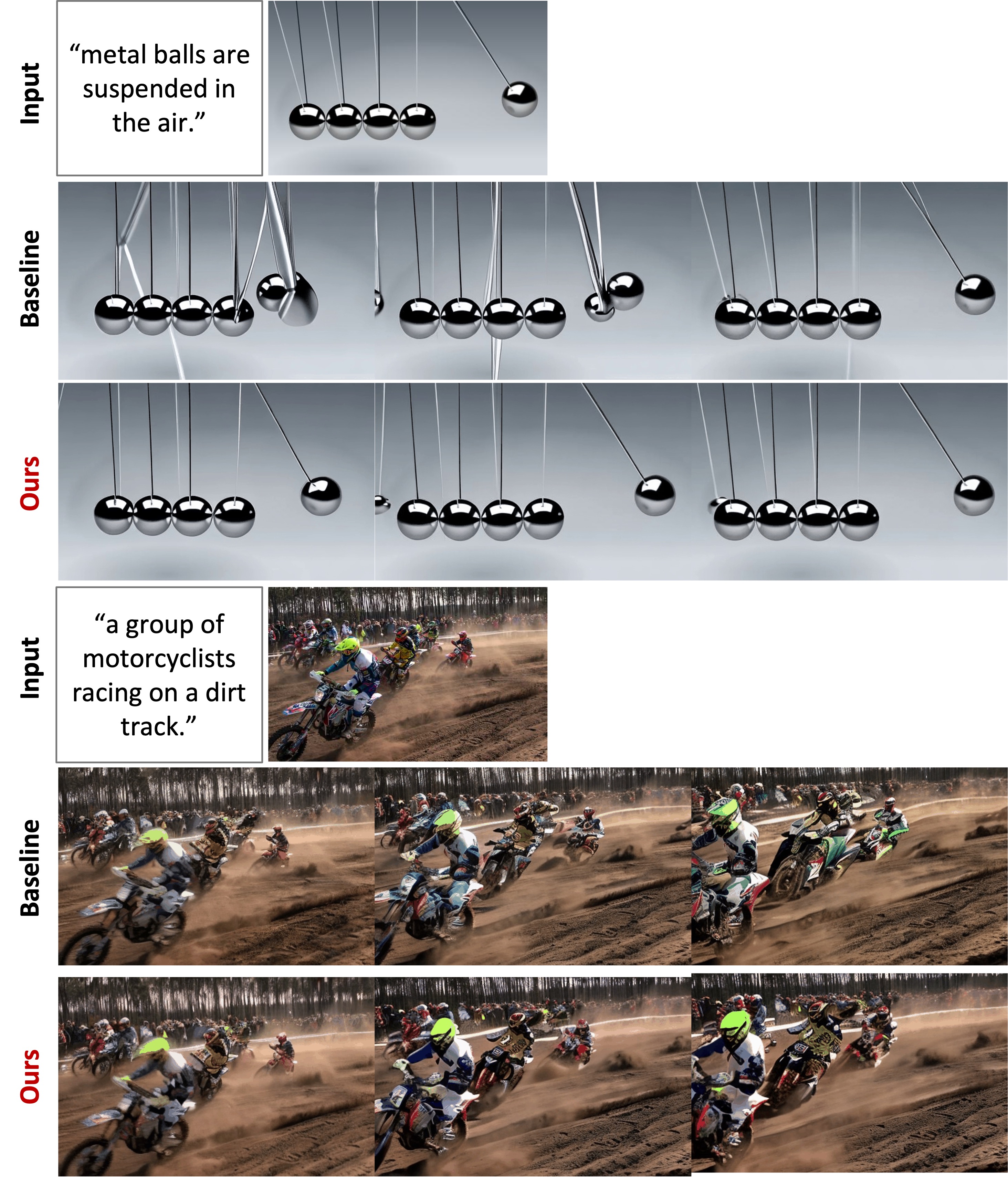}
    \caption{Additional results of DynamicCrafter~\cite{xing2025dynamicrafter}.}
    \vspace{-0.2cm}
    \label{fig:sup_add_dc}
\end{figure}

\subsection{Synergies with the Existing Method}

\begin{table}[t]
\centering
\resizebox{0.95\columnwidth}{!}{%
\begin{tabular}{c >{\columncolor{mylightgray}}c c c}
\hline
 & FI(1) + \textbf{Ours} & FI(2) & FI(4) \\
\hline
Subject Consistency & \underline{0.9669} & 0.9662 & \textbf{0.9711} \\
Background Consistency & \underline{0.9834} & 0.9828 & \textbf{0.9854} \\
Temporal Flickering & \underline{0.9579} & 0.9571 & \textbf{0.9672} \\
Motion Smoothness & \underline{0.9776} & 0.9771 & \textbf{0.9823} \\
Dynamic Degree & \underline{0.2850} & \textbf{0.2988} & 0.2600 \\
Overall Consistency & \textbf{0.2469} & \underline{0.2463} & 0.2424 \\
Image Quality & \textbf{0.6768} & \underline{0.6756} & 0.6435 \\
\hline
\end{tabular}%
}
\caption{Quantitative results of FreeInit and FreeInit combined with our method. FI denotes FreeInit, and the number in parentheses indicates the number of noise initialization steps performed. \textbf{Bold}: Best, \underline{Underline}: Second Best.}
\vspace{-0.2cm}
\label{tab:sp_freeinit}
\end{table}

We demonstrate how our approach can be combined with orthogonal methods to achieve enhanced performance.  For instance, while FreeInit ~\cite{wu2023freeinit} focuses on initializing noise, our method emphasizes guidance through prompt optimization. These complementary mechanisms allow the two approaches to work synergistically. In this section, we present the results obtained by using both methods together, highlighting their combined potential.

While FreeInit significantly improves temporal quality, it does so at the expense of overall video quality. Specifically, \autoref{tab:sp_freeinit} demonstrates that increasing the number of initialization steps leads to a sharp decline in metrics related to video generation quality, such as Overall Consistency and Image Quality. \autoref{fig:sp_freeinit} further reveal that, compared to the baseline, the videos generated with FreeInit often lose high-frequency details and exhibit noticeable saturation. However, by applying our method after a single noise initialization step, we observed an improvement in temporal quality while relatively minimizing the compromise in video quality, compared to the FreeInit approach where noise initialization is applied multiple times.


\subsection{Computational Cost}

\begin{wrapfigure}[4]{r}{0.55\linewidth}
    \centering
    \small
    \resizebox{0.95\linewidth}{!}{
        \begin{tabular}{ccc}
        \toprule
        Model & GPU Memory & Computing Time \\
        \midrule
        AnimateDiff & 11.38 GiB & 18.94 s/video \\
        \textbf{+Ours} & 35.60 GiB & 50.94 s/video \\
        \bottomrule
        \end{tabular}
    }
    \caption{Computational cost.}
    \label{tab:computation}
\end{wrapfigure}

We compute GPU memory usage and computing time, averaging over 100 generations.


\section{Additional Results}
\label{sup:results}
In this section, we provide additional result images to further demonstrate the performance and effectiveness of our approach across different models. First, we present additional results for DynamiCrafter~\cite{xing2025dynamicrafter}, an image-to-video model that takes prompts as input (\autoref{fig:sup_add_dc}). Furthermore, we provide results for AnimateDiff~\cite{guo2023animatediff}, Lavie~\cite{wang2023lavie}, and VideoCrafter2~\cite{chen2024videocrafter2}.

\begin{figure*}[t]
    \centering
    \includegraphics[width=1.0\linewidth]
    {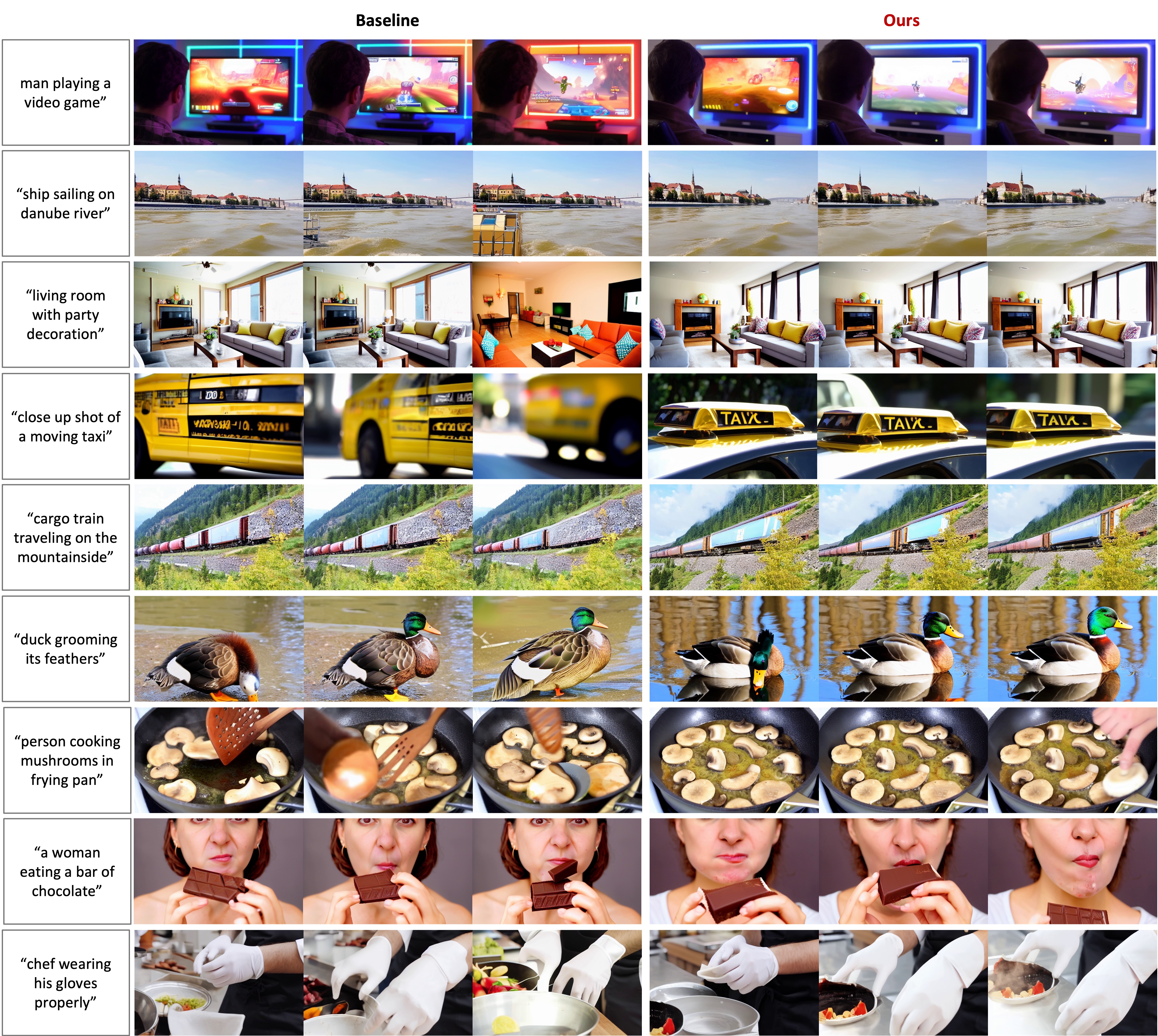}
    \caption{Additional results of Lavie~\cite{wang2023lavie}.}
    \vspace{-0.2cm}
    \label{fig:sup_add_lavie}
\end{figure*}

We observe that our method enhances temporal quality without significantly altering the generated content across these three models. Notably, in Lavie~\cite{wang2023lavie}, the last frame often deviates significantly from the preceding frames (see rows 3, 6, and 9 in \autoref{fig:sup_add_lavie}). Our approach effectively mitigates this issue to a large extent.

\begin{figure*}[t]
    \centering
    \includegraphics[width=1.0\linewidth]
    {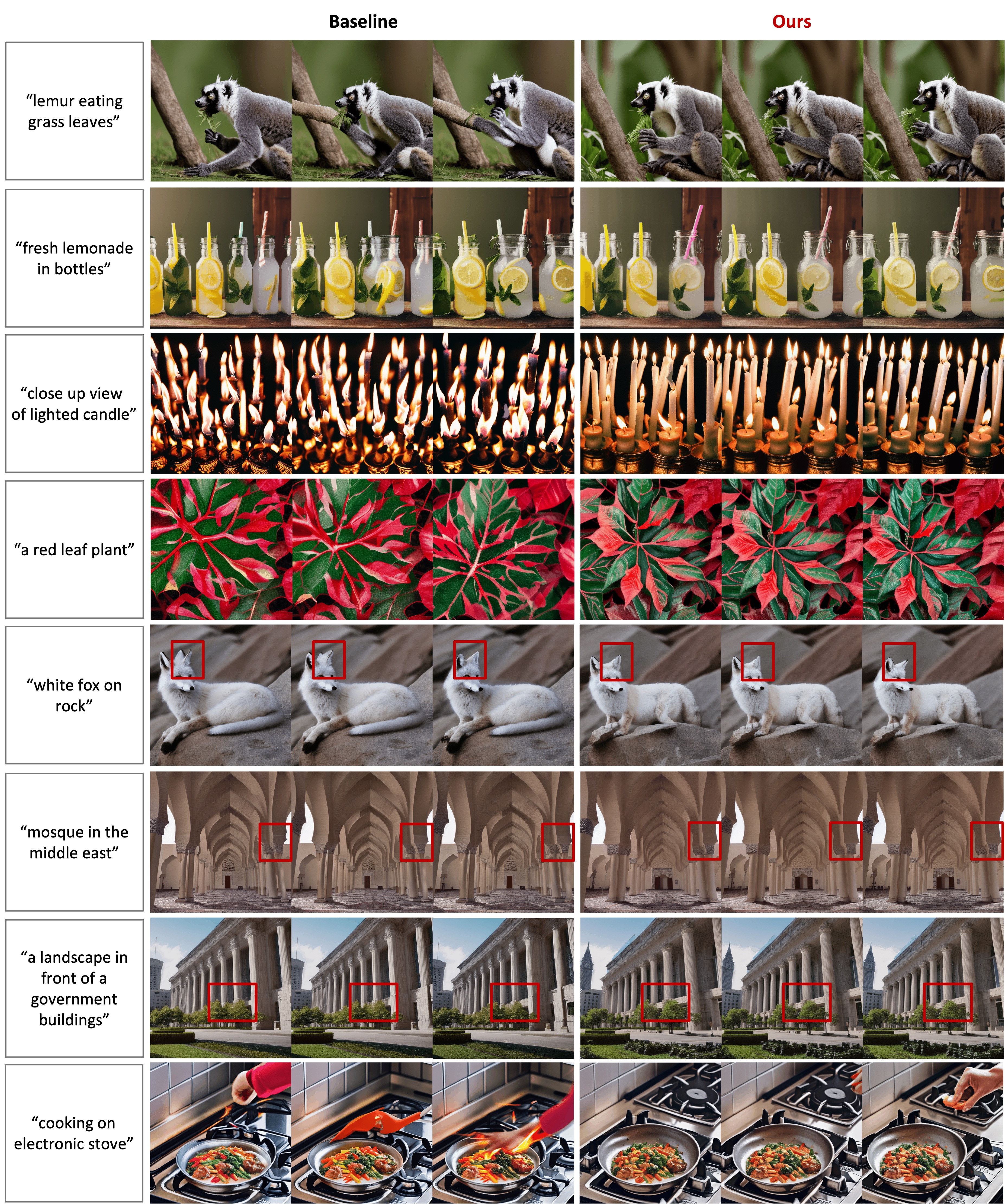}
    \caption{Additional results of AnimateDiff~\cite{guo2023animatediff}.}
    \vspace{-0.2cm}
    \label{fig:sup_add_ad}
\end{figure*}

\begin{figure*}[t]
    \centering
    \includegraphics[width=1.0\linewidth]
    {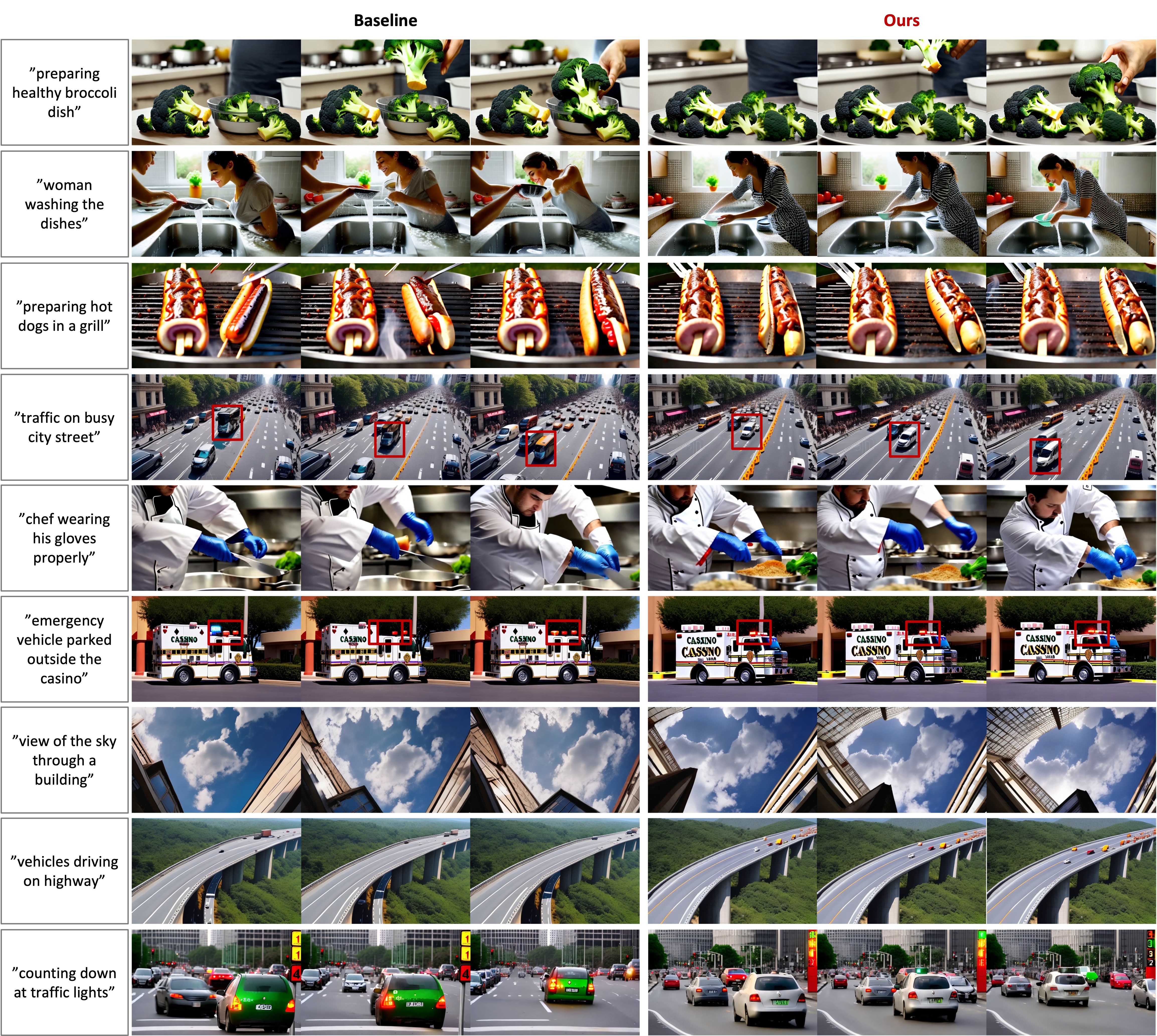}
    \caption{Additional results of VideoCrafter2~\cite{chen2024videocrafter2}.}
    \vspace{-0.2cm}
    \label{fig:sup_add_vc}
\end{figure*}

\end{document}